\documentclass[journal]{IEEEtran}

\usepackage{lineno,hyperref}
\usepackage[usenames, dvipsnames]{xcolor}
\usepackage{soul}
\usepackage{amsmath}
\usepackage{float}
\usepackage{subcaption}
\usepackage{adjustbox}
\modulolinenumbers[5]
\usepackage{tabularx,booktabs}
\usepackage{todonotes}

\definecolor{sand}{RGB}{255,255,187}
\definecolor{shale}{RGB}{142,142,139}

\usepackage{ulem}

\newcommand{\WOB}{\text{WOB}}
\newcommand{\RPM}{\text{RPM}}
\newcommand{\TRQ}{\text{TRQ}}

\newcommand{\ROP}{\text{ROP}}

\newcommand{\APR}{\text{APR}}
\newcommand{\SED}{\text{SED}}

\begin{document}

\title{Real-time  data-driven detection of the rock type alteration during a directional drilling}

\author{Evgenya Romanenkova,
        Alexey Zaytsev,
        Nikita Klyuchnikov,
        Arseniy Gruzdev,
        Ksenia Antipova,
        Leyla Ismailova,
        Evgeny Burnaev,
        Artyom Semenikhin,
        Vitaliy Koryabkin,
        Igor Simon,
        Dmitry Koroteev

\thanks{E. Romanenkova, A. Zaytsev, N. Klyuchnikov, K. Antipova, L. Ismailova, E.Burnaev D. Koroteev are with Skolkovo Institute of Science and Technology (Skoltech), 121205}
\thanks{A. Gruzdev and A. Semenikhin are with IBM East Europe/Asia, Moscow, 123112, Russia}
\thanks{I. Simon and V. Koryabkin are with Gazprom Neft Science \& Technology Center, St. Petersburg 19000, Russia}
\thanks{Manuscript submitted to IEEE Geoscience and Remote Sensing Letters}}

\markboth{IEEE Geoscience and Remote Sensing Letters,~Vol.~TODO, No.~TODO, TODO}%
{Instant data-driven detection of changes of the rock type during a directional drilling}

\maketitle

\begin{abstract}
During the directional drilling, a bit may sometimes go to a nonproductive rock layer due to the gap about 20m between the bit and high-fidelity rock type sensors. The only way to detect the lithotype changes in time is the usage of Measurements While Drilling (MWD) data. However, there are no general mathematical modeling approaches that both well reconstruct the rock type based on MWD data and correspond to specifics of the oil and gas industry. In this article, we present a data-driven procedure that utilizes MWD data for quick detection of changes in rock type. We propose the approach that combines traditional machine learning based on the solution of the rock type classification problem with change detection procedures rarely used before in Oil\&Gas industry. The data come from a newly developed oilfield in the north of western Siberia. 
The results suggest that we can detect a significant part of changes in rock type reducing the change detection delay from $20$ to $1.8$ meters and the number of false positive alarms from $43$ to $6$ per well.
\end{abstract}


\IEEEpeerreviewmaketitle

\section{Introduction}
The drilling goal is to construct a wellbore with maximum productivity: the significant part of a well should belong to a rock that produces oil. However, there is high uncertainty about the location of productive layers, as the preliminary geologic model uses observations at different locations.
Most of the information about the exact configuration of productive and not-productive layers in a well come during drilling.
Logging While Drilling (LWD) measurements allow close-to-online rock types detection.
However, there is about a $20$ m lag between the drilling bit and LWD sensors. 
This lag leads to wrong decisions and decreases the productive zone of a well. Another available information source is Measurements While Drilling (MWD).
We can measure them directly on a drilling bit.
But there are no established methods in physical modeling that allow real-time reliable rock types identification~\cite{klyuchnikov2018data}.

Increasing volumes of reliable data available in Oil \& Gas industry lead to the creation of data-driven models for rock type classification based on historical records from different wells \cite{klyuchnikov2018data,  kadkhodaie2010rock}. 
These articles accurately detect current rock type, but they are unable to provide precise rock type \textit{change detection} suitable for application in the industry.

\vspace{-10pt}
\begin{figure}[!h]
    \centering
    \includegraphics[width=0.9\linewidth]{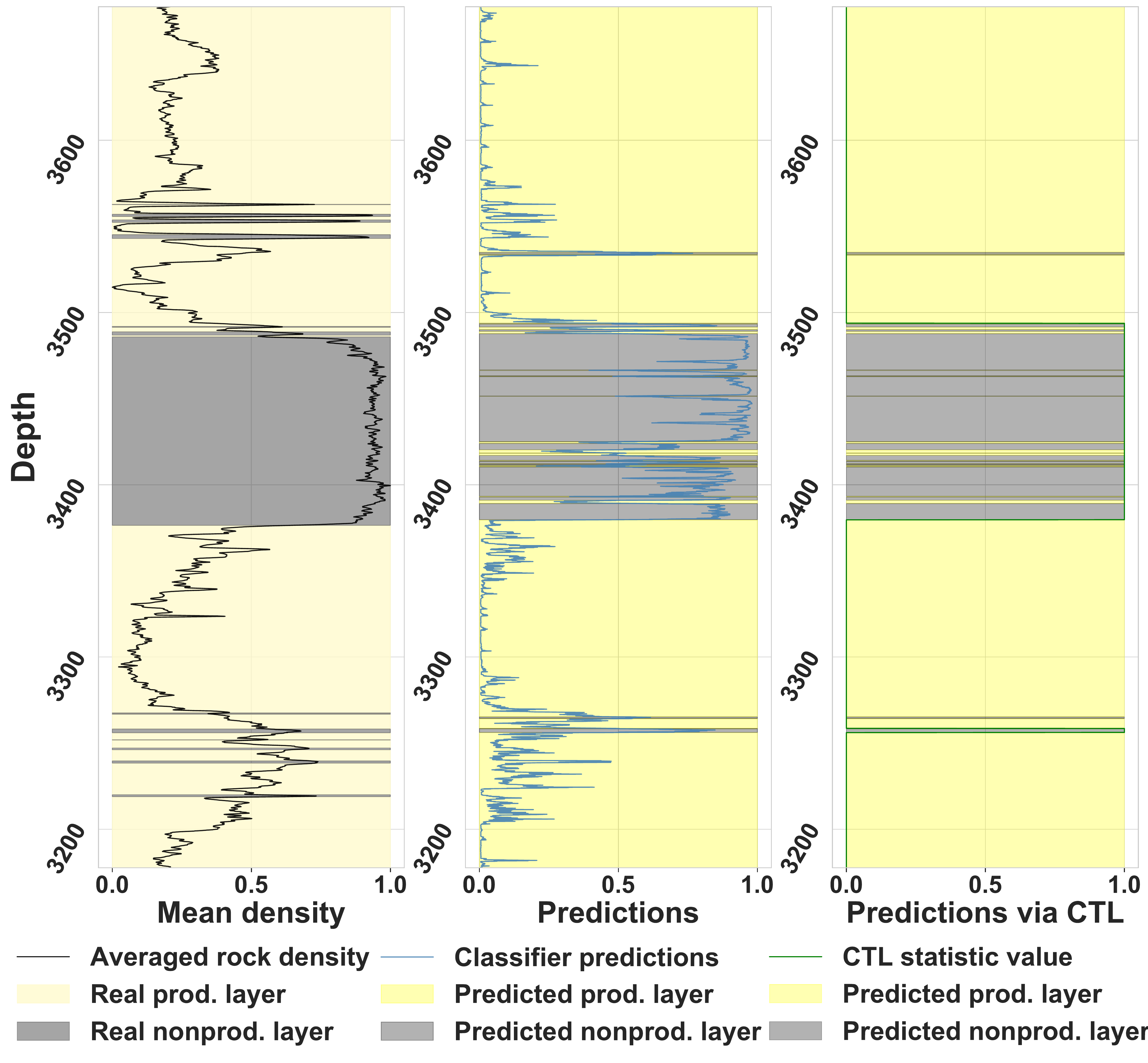}
    \caption{The proposed model in work. The true labels corresponding to the lithotype and density (left figure); Machine learning (ML) prediction for probabilities of rock types and corresponding labels (middle figure); result of cutting thin layers (CTL) on top of ML and corresponding labels (right figure). ML-CTL can detect most of the layers with a small number of false alarms.}
    \label{fig:teaser}
\end{figure}

To solve the problem at hand, we train two data-driven models: a machine learning (ML) classifier and a threshold-based change point detection algorithm. The real-time application of our models consists of two steps. After data preparation, by using the ML classifier, we predict the probabilities of lithotypes for each particular position of a drilling bit. Then applying the change-point detection algorithm to the probabilities, we signal about lithotypes changes.

The main contributions are the following:
\begin{itemize}
\item We collected a large dataset of MWD and LWD data for wells in the oilfield in the Siberia and explored data quality issues.
\item We developed a system that consists of three parts: data preprocessing, traditional ML approaches for rock type detection, and change detection on top of the ML approach. An example of the proposed system at work is in Figure \ref{fig:teaser}. 
\item We assessed the obtained solution: how strong the model performs in terms of its business value. We used mean delay detection, the specific metric of the change detection problems, and a more industry-related Accuracy N~\cite{simon2018determination} as well as the confusion matrix. 
\end{itemize}

\subsection{Prior art of the MWD-based rock-type identification} 

\subsubsection{Physical Modeling for the rock-type identification}

Physical models are based on the physical equations (typically mass and energy balances) governing the behavior of the system under analysis. 
A review~\cite{Sugiura2015} can be considered as 
the most accurate description of the state-of-the-art 
in the modeling of drilling systems for automation and control, adaptive modeling for downhole drilling systems, and actual industry tasks.

There are a lot of more specific papers. 
\cite{Downton2012} examines the modeling of different aspects of drilling and focuses on the possibility of bringing these models into unified control systems to fully automate the entire process.
Paper \cite{Cayeux2014} analyzes the sensor equipment on the drilling rig and the issues of its layout for modeling of the drilling process.

Simplification of the drilling process leads to analytical formulae \cite{Detournay1992}. 
Estimation of the coefficients for all rock and bit types is possible for a big enough data sample, allowing direct rock type identification.

However, at the moment, physical modeling methods are not mature enough for application in rock type identification due to either high computational complexity, limitations on the data quality for physics-based models, or low accuracy of more empirical ones~\cite{Sugiura2015, Downton2012}. 

\subsubsection{Data-driven approaches for the rock-type identification}

Some papers consider data-driven approaches for rock type detection using MWD data.
Paper \cite{zhou2010hybrid} examines an application of supervised and unsupervised learning for on-bit rock typing with MWD data. 
The list of features used in the data-driven model consists of a rate of penetration \ROP{}, a weight on bit \WOB{}, and top drive torque \TRQ{} as the input parameters for the data-driven model. 
Authors propose to use adjusted penetration rate 
$\APR{} = \ROP{} / ({\WOB{}}\sqrt{\TRQ{}})$. 

They conclude that the combination of the supervised classification based on Gaussian processes and a clustering leads to the best performance.
Another widely used feature intended to capture the rock type effects is the Specific Energy of Drilling that depends on $\WOB{},  \RPM{}, \TRQ{}, \ROP{}$ and a cross-section area of the wellbore. 
Paper~\cite{zhou2011adaptive} suggests that unsupervised learning accompanied by minimization of \SED{} can lead to the penetration rate increase. 
Another effort on the penetration rate optimization is presented in \cite{hegde2017use}. 
The authors used Random Forest algorithm to build a model linking the penetration rate with \WOB{}, rotation speed \RPM{}, drilling mud rate, and unconfined rock strength. 
The usage of the model increased the penetration rate for up to $12\%$ for the wells close to ones used during training. 

The paper \cite{labelle2001lithological} describes the artificial neural networks (ANNs) application for rock typing at drilling. 
MWD-like measurements as input for ANNs and decision trees provide a model with a small $4.5\%$ classification error for a complete enough set of mechanical features is available \cite{simon2018determination}, \cite{klyuchnikov2018data}. In the paper \cite{kadkhodaie2010rock}, the authors examined boosting and ANNs to predict rock-type using MWD data. However, they used a grid of close vertical wells with small depths. 

We build on these ideas to construct a model for the Oil \& Gas industry and a diverse set of oil fields. The area of the model application should include horizontal wells since directional drilling is the only way to deal with many complex conditions. 

\subsubsection{Change detection overview}

Time series data are sequences of measurements at consecutive moments which describe the behavior of some system. 
A system can change because of external events. Understanding the causes of variation and predicting the moment of change is the aim of the change detection. Most of the change detection approaches take ideas from statistics and ML and apply them to solve problems in diverse areas like medicine, industry, image analysis, and manufacturing~\cite{aminikhanghahi2017survey}.

The statistics-based approaches involve the usage of various statistics such as cumulative sums, Shiryaev-Roberts statistics, a prior distribution statistic, as well as an application of filters and smoothing, see survey \cite{aminikhanghahi2017survey}.
For example, in \cite{Degradation2016} the authors mixed Kalman filter, the cumulative sum, and the exponentially weighted moving average to detect degradation of software-intensive systems in real-time. 
The ML-based approach is presented, e.g. in review \cite{wang2018supervised}, where authors offered ANN methods for speech separation.

Results on change detection in Oil\&Gas industry are scarce.
Most of the articles consider the detection of abnormal behavior for time series: 
in \cite{marti2015anomaly} authors used time series segmentation on top of a kernel method for anomaly detection for oil platform turbomachinery;
in \cite{ide2016sparse} authors apply Gaussian Markov random fields-based approach to detection of anomalies during offshore oil production;
in \cite{yan2015accurate}, authors applied anomaly detection for the health monitoring of gas turbine combustors. 

In contrast to the common anomaly detection, here we identify switching between two behaviours of a bit: drilling in the oil-bearing and in non-productive shale layers. 
This problem is the detection of multiple consecutive changes.

\section{Data overview}

The considered oil and gas condensate field is located within the Yamal Peninsula.
It is the largest field under development in Siberia northwest. 
We consider $57$ wells for the most productive formation of Lower Cretaceous. 
The mean well length is $\sim 3800$ m, and the formation depth is about $1800$ m. 
In this work, we operate only horizontal wells.

The initial data included MWD, LWD data from downhole sensors.
In turn, MWD data contains the following parameters: \WOB{}, rotary speed, \TRQ{}, input and output flow rate, standpipe pressure, \ROP{}, and hook land.

The rock-types are the result of the petrophysical interpretation of LWD measurements, that consist of natural gamma radiation, apparent resistivity, polarization resistance.
Those rock-types were used as the target variable during the construction of data-driven models and assessment of these models.

\section{Methods}

\subsection{Data preprocessing}

We preprocess the data and represent it as a sequence of measurements: we divide each well into depth intervals of size $0.1$ meters, and so any moment $t$ corresponds to some particular depth.

In our case, it was essential to handle missing data and significant class imbalance: 
there are only $16.55\%$ of shales and hard-rocks in the available data and $83.45\%$ of sands.  
We deal with the missing values in the following way: 
we fill the intervals of constant lithotype with at least one value available using the mean of known values, if there were no available values, we use the latest given preceding value. This method allowed us to work with 99.8\% of the initial data.

\subsection{General approach to problem}

We need to distinguish two types of rock.
Thus, first, we solve a standard classification problem, i.e. predict lithotype at each moment, and then use a change detection approach on top of the classifier predictions: 
\begin{enumerate}
    \item Get predictions for rock types using Gradient Boosted Decision Trees over MWD data (see sec. ~\ref{sec:ml_approaches}).
    \item Get the updated labeling by aggregating these predictions with statistics (see sec. ~\ref{sec:change_detection}).
\end{enumerate}
Then we can evaluate quality metrics for obtained labeling (see sec. ~\ref{sec:quality_metrics}).

\subsection{Machine learning classification approaches}\label{sec:ml_approaches}

\paragraph{Decision trees and gradient boosting}

One of the most widely used approaches for classifiers construction is Ensembles of decision trees~\cite{fernandez2014we}. 
For a constructed decision tree, we proceed through it according to the values of input variables for input object until it reaches a leaf;
in a leaf the basic classifier returns the probabilities to belong to classes.
In an ensemble we combine weighted basic
decision tree classifiers.
Ensembles of decision trees are fast to construct, robust to over-fitting, handle missing values and outliers, and provide competitive performance~\cite{fernandez2014we}.

We use Gradient boosting algorithm for the construction of ensembles of decision trees~\cite{chen2016xgboost}. 
The algorithm has the following main hyperparameters: 
the number of trees in an ensemble, the maximum depth of each tree, 
the share of features used in each tree, the share of samples used for training of each tree,
and the learning rate. 

\paragraph{Artificial Neural Networks}

An alternative data-driven approach is ANNs~\cite{goodfellow2016deep}.
They are more sensitive to the quality and size of input data and hyperparameters.
However, they are widely used while dealing with specific data structures like images or time series~\cite{AHMAD201777}.

We tried classic Feedforward ANNs \cite{Hornik:1989:MFN:70405.70408}, as well as recurrent ANNs \cite{Hochreiter:1997:LSM:1246443.1246450}, End-to-End Deep ANNs and domain adaptation approaches~(see Chapter 10 in \cite{goodfellow2016deep}). 
Results were unstable due to the properties of ANNs and data quality issues. 

\subsection{Change detection approaches}\label{sec:change_detection}

\paragraph{Statistical approaches} Change detection approaches process data iteratively to detect the moment when data properties change. In our case, we consider the classifier outputs as observations, and the change point occurs when the drilling bit enters a layer of another lithotype. 

We apply three different change detection statistics to the output probabilities of the classifier: cumulative sums, Shiryaev-Roberts statistics, and posterior probabilities statistics \cite{aminikhanghahi2017survey}. For all statistics, the depth of lithotype change $\tau_s$ is the moment $t$ when the statistics value $S_n$ exceeds a threshold $w$: $\tau_s = \inf \{t \geq 0: S_t \geq w \}$. Two thresholds are the main hyperparameters of the techniques: the first one is related to alteration between oil-bearing interval and a non-productive shale layer, and the second one is used to detect a change in the reverse direction. 
The posterior probabilities statistic also has a hyperparameter $p$ that identifies the prior distribution of the moment of change.

\paragraph{Dropping thin layers} 
Our drilling data contain a lot of thin layers that are hard to detect. Even if we could identify the beginning of a thin layer using statistics, we can hardly accumulate enough information to detect its end.

The idea of dropping thin layers is to replace the predicted label of a detected thin layer with that of the previous one. 
We suppose that this method can improve quality by reducing the number of false alarms. 
The only hyperparameter of this method is an upper bound of the layer size to drop $w$. 

While it seems that cutting thin layer is a heuristic, it has the origin similar to that of statistical approaches as it utilizes the history of likelihood values. 
Actually, we can represent dropping a thin layer in two steps. First, based on the likelihood for a depth $(t + 1)$ we calculate statistic $S_{t + 1}$: if the classifier output $l_t < l_0$ for some threshold $l_0$, then $S_{t + 1} = 0$, otherwise $S_{t + 1} = S_t + 1$.
\[
S_{t + 1} = 
 \left\{
     \begin{gathered} 
         S_t + 1, \ l_{t+1} \geq l_0, \hfill \\ 
         0, \ l_{t+1} < l_0. \hfill 
         \\ 
     \end{gathered} 
 \right.
 \]
Second, we compare the obtained statistics with $w$. 
If $S_t > w$, we set all the previously predicted labels $y_{t - w}, \ldots , y_{t}$ to the value of the label $y_{t - w - 1}$.

\subsection{Quality metrics}
\label{sec:quality_metrics}
Although accuracy, ROC AUC, PR AUC, precision and recall are widely used in ML
(see \cite{grau2015prroc,gurina2019application} for the definitions), they are weak at measuring the quality of a change detection approach. 
Thus, we choose more suitable metrics: 
Accuracy L, Accuracy N~\cite{simon2018determination}, mean delay of change detection, and some characteristics of a confusion matrix for change detection. 
If the distance is bigger than $20$ m, we detect the lithotype change via LWD, so we take into account only changes detection during the first $20$ m after the real change.

According to our observations, the elongation of a drilling bit is approximately $1.5$ m. 
Therefore, a drilling bit reacts to lithotype changing after $1.5$ m delay due to its deformation. This leads us to Accuracy L, an analogue of the usual accuracy that ignores neighborhood type of rock changes in the window of $1.5$ m at the ends of lithotype layers.  
To calculate Accuracy N, we use the following procedure:
\begin{enumerate}
\item Exclude from data neighborhood of rock type changes of size $1.5$ m at ends of lithotype layers.
\item Split data into intervals according to the actual and predicted changes of rock type.
\item For each interval, identify whether the predicted rock type coincides with the actual one.
\item Calculate the percentage of correct predictions.
\end{enumerate}

\begin{figure}[t]
\centering
\includegraphics[width=0.44\textwidth]{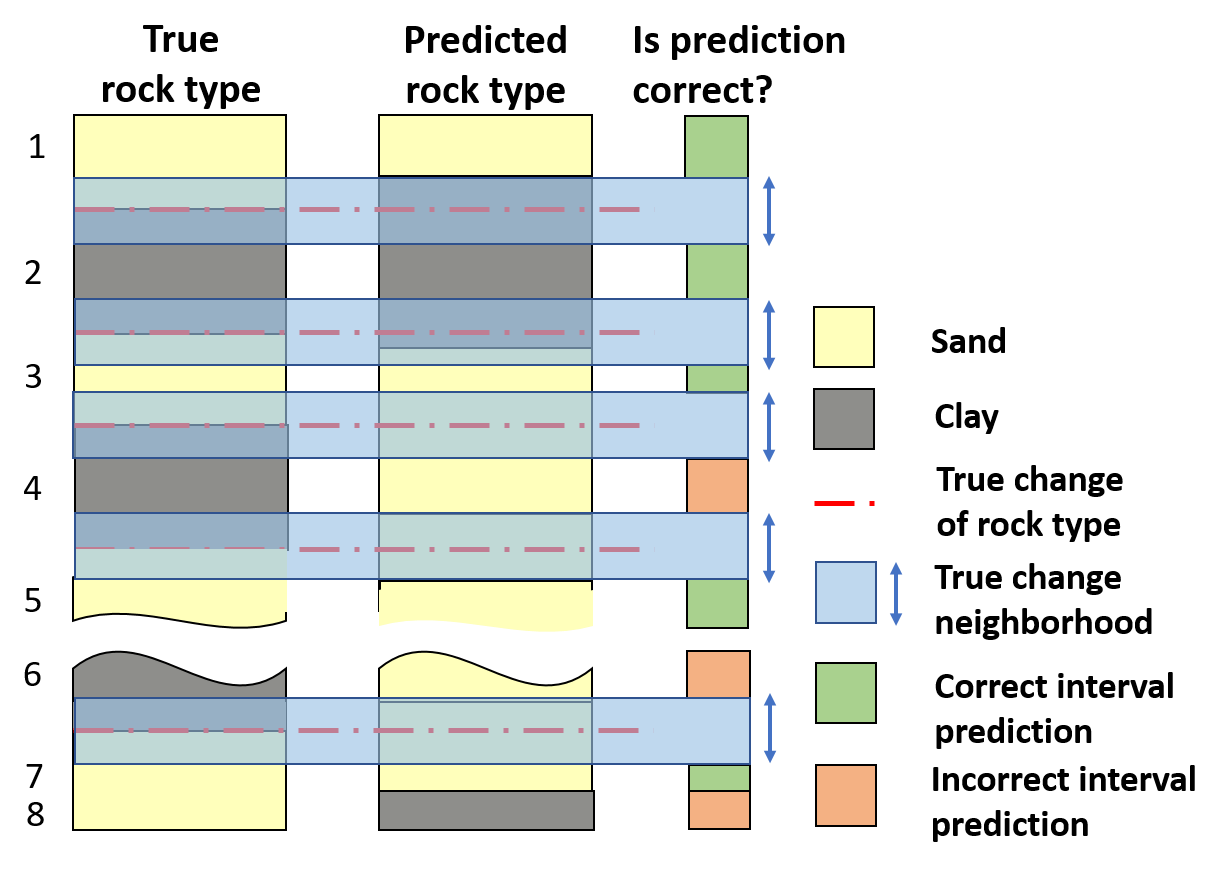}
\caption{Example of Accuracy N calculation. Predicted rock type is correct for 5 intervals (1, 2, 3, 5 and 7) and incorrect for 3 other intervals (4, 6 and 8), Accuracy N is $\frac{5}{5 + 3} = 0.625$.}
\label{fig:acc_n_example}
\end{figure}

Evaluation of Accuracy N is illustrated by~Figure~\ref{fig:acc_n_example}.

We consider three natural quality measures of change detection:
\begin{itemize}
    \item delay detection characterizes the difference in time between the actual and detected change if it is captured \cite{Shiryaev2009};
    \item the percentage of changes correctly predicted within 20 meters; 
    \item the number of True Positive alarms that shows the number of lithotype changes detected in the next $3$ m after they appear;
    \item the number of False Positive alarms that shows the number of predicted changes while they don't appear in the previous $3$ m.
\end{itemize} 
The first and the last metrics are antagonistic: a model with large sensitivity detects changes early but raises a lot of False alarms. Similarly, using a model with extremely low sensitivity, we tend to have no False positives errors, but also miss all actual changes. A good model should meet a trade-off between quick change detection and rare false alarms errors. 
To measure the models quality, we use leave-one-well-out cross-validation:
we exclude wells from data one-by-one, train the model using all other wells, and get predictions for a hold-out well.
Then we aggregate results taking medians of metrics over all wells.
This procedure allows us to avoid over-fitting, that occurs when we train and test the model using the data from the same well: according to ~\cite{klyuchnikov2018data}, the data inside one well are significantly closer than the data from two separate wells. 
We estimate hyperparameters of methods by grid search maximization of Accuracy N scores obtained via cross-validation.

\section{Results}

\subsection{Machine learning approach}
We compare different ML models: boostings of decision trees (Gradient Boosting (GB) from sklearn \cite{sklearn}, LogitBoost, XGBoost) and ANNs. 
The hyperparameters for boosting are the following:
\begin{itemize}
    \item \textit{number of decision trees:} 200;
    \item \textit{max depth for each tree:} 3;
    \item \textit{learning rate:} 0.05. 
\end{itemize}

Besides, we analyze ANNs with a various number of layers, ReLu nonlinearities, and $20$ neurons at each hidden layer.

The result of the comparison presented in Table~\ref{tab:ml_methods}. Boosting models work better than ANNs. Besides, XGBoost has slightly higher aggregated metrics (ROC AUC, PR AUC) and the important business metric Accuracy N.

\begin{table*}[h!]
    \centering
    \begin{tabular}{l|ccc|ccc}
    \hline
     &  \multicolumn{3}{c}{Ensembles of decision trees} & \multicolumn{3}{|c}{Neural networks (ANNs)}\\
     \hline
     Metrics values & GB from sklearn & LogitBoost & XGBoost & 1 layer NN & 2 layers NN & 3 layers NN\\
     \hline
     ROC AUC & 0.914 $\pm$ 0.060 & 0.911 $\pm$ 0.074 & \textbf{0.919 $\pm$ 0.060} & 0.891 $\pm$ 0.121 & 0.909 $\pm$ 0.129 & 0.904 $\pm$ 0.118 \\ 
     PR AUC & 0.699 $\pm$ 0.200 & 0.689 $\pm$ 0.201 & \textbf{0.705 $\pm$ 0.200} & 0.648 $\pm$ 0.220 & 0.698 $\pm$ 0.224 & 0.677 $\pm$ 0.218 \\
     Acc. L & 0.961 $\pm$ 0.073 & 0.962 $\pm$ 0.072 & \textbf{0.963 $\pm$ 0.073} & 0.953 $\pm$ 0.136 & 0.954 $\pm$ 0.108 & 0.945 $\pm$ 0.095 \\
     Acc. N & 0.545 $\pm$ 0.063 & 0.549 $\pm$ 0.064 & \textbf{0.551 $\pm$ 0.062} & 0.529 $\pm$ 0.043 & 0.535 $\pm$ 0.048 & 0.535 $\pm$ 0.046 \\ 
     Accuracy & 0.921 $\pm$ 0.073 & \textbf{0.924 $\pm$ 0.072} & 0.923 $\pm$ 0.073 & 0.909 $\pm$ 0.125 & 0.910 $\pm$ 0.103 & 0.905 $\pm$ 0.092 \\ 
     Precision & \textbf{0.806 $\pm$ 0.225} & 0.786 $\pm$ 0.222 & 0.779 $\pm$ 0.219 & 0.747 $\pm$ 0.250 & 0.755 $\pm$ 0.263 & 0.725 $\pm$ 0.286 \\ 
     Recall & \textbf{0.555 $\pm$ 0.221} & 0.513 $\pm$ 0.220 & 0.545 $\pm$ 0.220 & 0.450 $\pm$ 0.216 & 0.540 $\pm$ 0.248 & 0.524 $\pm$ 0.265 \\ 
     \hline
    \end{tabular}
    \caption{We compare the diverse approaches for the building ML model. All metrics are the larger, the better. XGBoost model is the best choice.}
    \label{tab:ml_methods}
\end{table*}

\vspace{-10pt}

\subsection{Statistical change detection}
We compare statistics and dropping thin layers. First, we choose the following hyperparameters for different statistics by maximizing the most business-related metric Accuracy N:

\begin{itemize}
    \item \textit{cutting thin layer}: length of the layer for cutting --- 15;
    \item \textit{cumulative sums}: first threshold --- 25, second --- 50;
    \item \textit{Shiryaev-Roberts statistics}: first and second thresholds --- $1158 \cdot 10^9$;
    \item \textit{prior distribution statistics}: first threshold --- $8 \cdot 10^5$, second threshold --- $7 \cdot 10^5$, probability --- $0.1$.
\end{itemize}

The considered metrics for these hyperparameters are in Table~ \ref{tab:statistic_metrics}. 
The drop of thin layers provides higher accuracy than other change detection statistics due to the smaller number of hyperparameters and the absence of additional probabilistic assumptions easily violated for drilling data.

\begin{table*}
    \centering
    \begin{tabular}{lcccccc}
        \hline
        \textbf{Used method} & \textbf{Acc. L} & \textbf{Acc. N} & 
        \begin{tabular}{c}
            \textbf{Mean detection}\\
            \textbf{delay}
        \end{tabular} &
        \begin{tabular}{c}
            \textbf{\% of identified}\\
            \textbf{change in 20 m}\\
        \end{tabular}
        & 
        \begin{tabular}{c}
            \textbf{True positive}\\
            \textbf{number}
        \end{tabular}
        & 
        \begin{tabular}{c}
            \textbf{False positive}\\
            \textbf{number}
        \end{tabular}\\
        \hline
        Only classification model & 0.9635 & 0.5511 & \textbf{0.3} & 0.854 & \textbf{25} & 43\\ 
        Cutting thin layers & 0.9655 & \textbf{0.6579} & 1.8 & \textbf{0.912} & 12 & \textbf{6}\\
        Cumulative sums & 0.9684 & 0.6551 & 4.0 & 0.881 & 8 & 6\\
        Shiryaev-Roberts statistics & \textbf{0.9701} & 0.6400 & 3.7 & 0.878 & 6  & 7\\ 
        Priory distribution statistics & 0.9685 & 0.6207 & 3.5 & 0.849 & 6 & 12\\
        \hline
    \end{tabular}
    \caption{We consider the median values of the metrics for $57$ wells. Cutting  thin  layers is not the best approach according to all metrics, but it improves Accuracy N and number of False Positives.}
    \label{tab:statistic_metrics}
\end{table*}
\vspace{-10pt}

\subsection{Analysis of the best model}

\begin{figure}[t]
    \centering
    \includegraphics[width=0.45\textwidth]{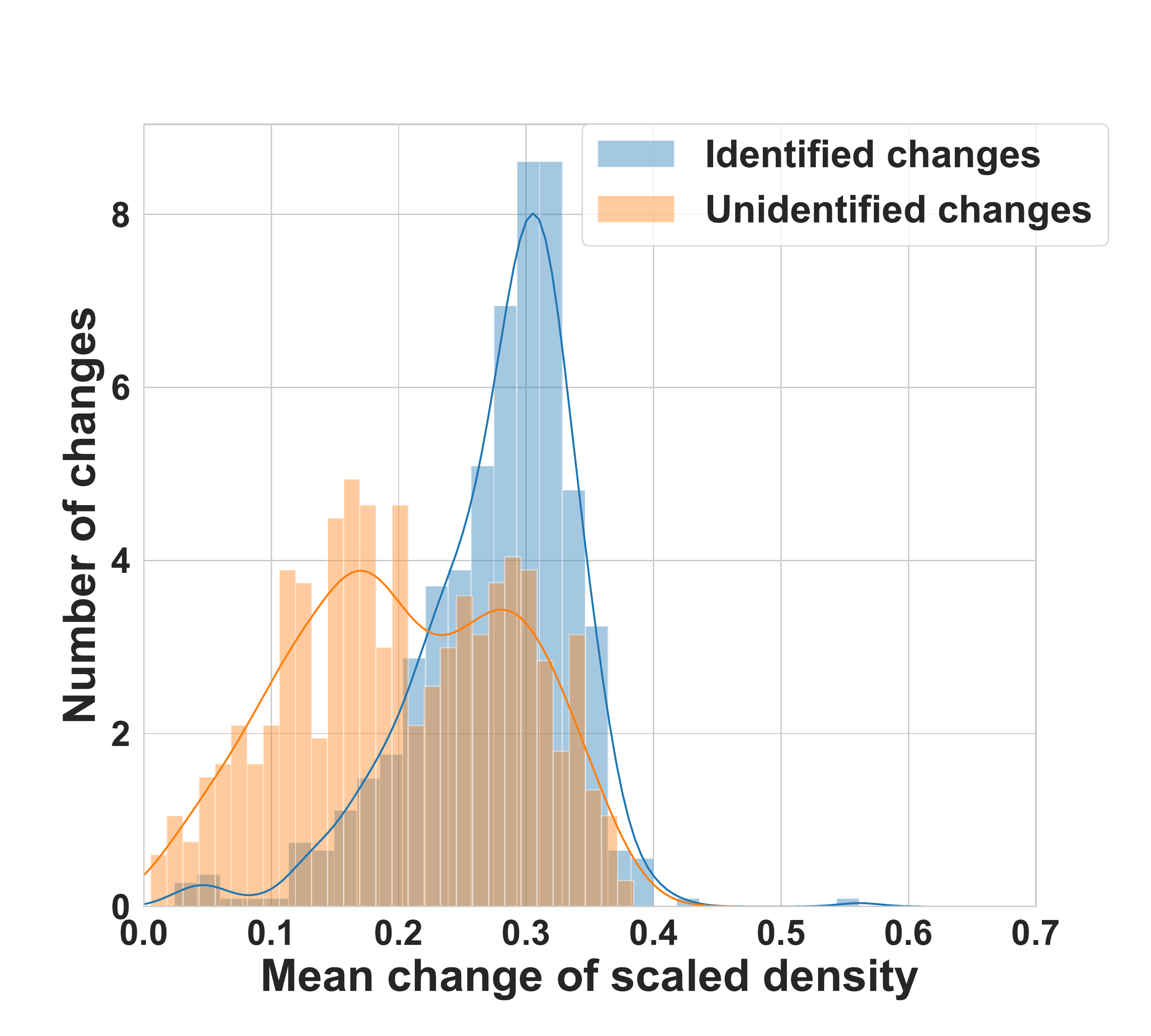}
    \caption{Histograms of density differences for identified and unidentified changes. Errors in detection can appear because the lithotype change in wellbores can occur gradually, so there are areas with mixed rock types.}
    \label{fig:hist_distplot_pred}
\end{figure}

Our hypothesis is that it is harder to detect a change in a rock type if rock types of neighbour intervals have similar properties.
To identify similarity, we measure the absolute difference in mean LWD densities for these layers for each change in our sample.
Histograms for similarities for identified and unidentified changes are in Figure~\ref{fig:hist_distplot_pred}:
our approach is better at the identification of changes if two neighbour layers differ significantly.

\vspace{-10pt}

\section{Conclusion}

We proposed the model for the rock type detection based on MWD data. Our system consists of three components: data preprocessing; 
classification of rock types using ML Gradient Boosted Decision Trees; aggregation of classifier predictions to get change detection using cutting of thin layers. 
The system identifies 93$\%$ changes within $20$ m range, has only about $7$ false alarms per well, and has $2.59$ m mean change detection delay. 

Moreover, according to our experiments, the main problem with change detection procedures was due to small differences between neighbour layers; thus, the detection of those changes would be difficult for a human as well.
\vspace{-5pt}

\bibliography{mybibfile}

\end{document}